\documentclass[]{bytedance_seed}

\usepackage[toc,page,header]{appendix}

\usepackage{minitoc}
\usepackage{algorithm}
\usepackage{algpseudocode}
\usepackage{enumitem}
\usepackage{xcolor}
\usepackage{amsfonts}
\usepackage{amssymb}

\title{VAPO: Efficient and Reliable Reinforcement Learning for Advanced Reasoning Tasks }

\affiliation[]{ByteDance Seed}
\contribution{Full author list in Contributions}

\abstract{
We present VAPO, \textbf{V}alue-model-based \textbf{A}ugmented Proximal \textbf{P}olicy \textbf{O}ptimization framework for reasoning models., a novel framework tailored for reasoning models within the value-model-based paradigm. Benchmarked  the AIME 2024 dataset, VAPO, built on the Qwen 32B pre-trained model, attains a state-of-the-art score of $\mathbf{60.4}$. In direct comparison under identical experimental settings, VAPO outperforms the previously reported results of DeepSeek-R1-Zero-Qwen-32B and DAPO by more than 10 points. The training process of VAPO stands out for its stability and efficiency. It reaches state-of-the-art performance within a mere 5,000 steps. Moreover, across multiple independent runs, no training crashes occur, underscoring its reliability. This research delves into long chain-of-thought (long-CoT) reasoning using a value-model-based reinforcement learning framework. We pinpoint three key challenges that plague value-model-based methods: value model bias, the presence of heterogeneous sequence lengths, and the sparsity of reward signals. Through systematic design, VAPO offers an integrated solution that effectively alleviates these challenges, enabling enhanced performance in long-CoT reasoning tasks.
}

\date{\today}
\correspondence{Yu Yue at \email{yueyu@bytedance.com}}

\begin{document}
\maketitle


\vspace{-15pt}
\begin{figure}[h]
    \centering
    \includegraphics[width=0.85\linewidth]{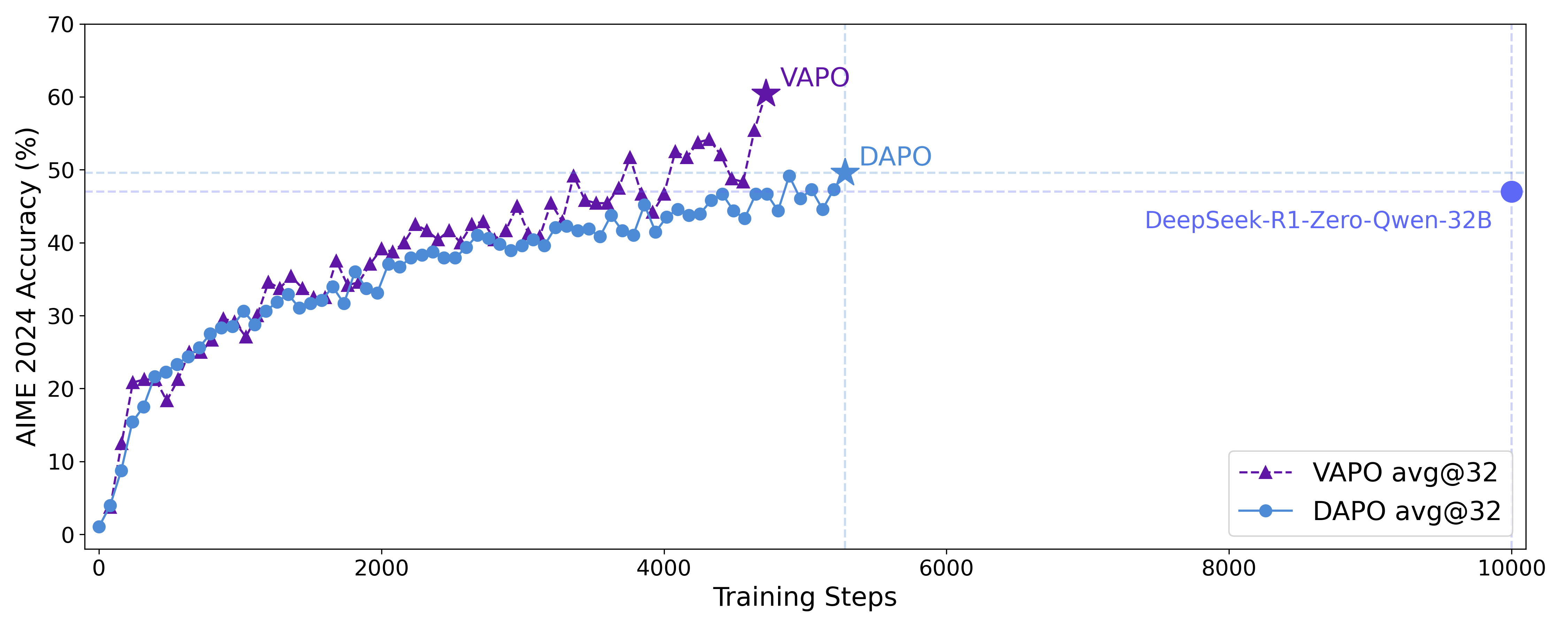}
    \caption{AIME 2024 scores of \textbf{VAPO} on the Qwen2.5-32B base model, demonstrates significant superiority over the previous state-of-the-art (SOTA) method DAPO, achieving this with notably fewer training steps. The x-axis denotes the gradient update steps.
    }
    \label{fig:front}
\end{figure}

\section{Introduction}



Reasoning models~\cite{gemini-thinking,qwq,k1.5} such as OpenAI O1 \citep{o1} and DeepSeek R1 \citep{deepseekai2025deepseekr1incentivizingreasoningcapability} have significantly advanced artificial intelligence by exhibiting remarkable performance in complex tasks such as mathematical reasoning, which demand step-by-step analysis and problem-solving through long chain-of-thought (CoT)~\cite{cot_2022} at test time.
Reinforcement learning (RL) plays a pivotal role in the success of these models~\cite{k1.5,areal,dapo,shao2024deepseekmath,hu2025reinforce++,rloo,ahmadian2024basicsrevisitingreinforcestyle,sutton1998rl}. It gradually enhances the model's performance by continuously exploring reasoning paths toward correct answers on verifiable problems, achieving unprecedented reasoning capabilities.

In the Large Language Models (LLM)~\cite{gpt3,gpt4,dsv3,grok,team2023gemini,claude35sonnet,chowdhery2023palm} RL training, value-model-free methods like GRPO~\cite{shao2024deepseekmath} and DAPO~\cite{dapo} have demonstrated remarkable effectiveness. These approaches eliminate the computational overhead of learning a value model, instead computing advantage solely based on the final reward of the entire trajectory. The trajectory-level advantage is then directly assigned as the token-level advantage for each position in the sequence. When training a reliable value model is particularly challenging, value-model-free methods deliver an accurate and stable baseline for advantage calculation by averaging the rewards across multiple trajectories within a group. This group-based reward aggregation mitigates the need for explicit value estimation, which often suffers from instability in complex tasks. Consequently, value-model-free methods have gained significant traction in addressing difficult problems such as long-CoT reasoning, with substantial research efforts focused on optimizing their frameworks.

Despite the notable success achieved by the value-model-free methods, we argue that value-model-based approaches possess a higher performance ceiling if the challenges in training value models can be addressed. First, value models enable more precise credit assignment by accurately tracing the impact of each action on subsequent returns, facilitating finer-grained optimization~\cite{ppo}. This is particularly critical for complex reasoning tasks, where subtle errors in individual steps often lead to catastrophic failures, and it remains challenging for model optimizing under value-model-free frameworks~\cite{vc-ppo}. Secondly, in contrast to the advantage estimates derived from Monte Carlo methods in value-model-free approaches, value models can provide lower-variance value estimates for each token, thereby enhancing training stability. Furthermore, a well-trained value model exhibits inherent generalization capabilities, enabling more efficient utilization of samples encountered during online exploration. This significantly elevates the optimization ceiling of reinforcement learning algorithms. Consequently, despite the formidable challenges in training value models for complex problems, the potential benefits of overcoming these difficulties are substantial.

However, training a perfect value model in Long COT tasks presents significant challenges.
First, learning a low-bias value model is non-trivial given the long trajectory and the instability of learning value in a bootstrapped way. Second, handling both short and long responses simultaneously is also challenging, as they might exhibit very distinct preferences towards the bias-variance trade-off during optimization. Last but not least, the sparsity of the reward signal from verifiers is further exacerbated by the long CoT pattern, which intrinsically requires better mechanisms to balance exploration and exploitation. To address the aforementioned challenges and fully unleash the potential of value-model-based methods in reasoning tasks, we present \textbf{V}alue \textbf{A}ugmented proximal \textbf{P}olicy \textbf{O}ptimization (\textbf{VAPO}), a value-model-based RL training framework. VAPO draws inspiration from prior research works such as VC-PPO~\cite{vc-ppo} and DAPO~\cite{dapo}, and further extends their concepts.

We summarize our key contributions as follows:
\begin{enumerate}
\item We introduce VAPO, the first value-model-based RL training framework to outperform value-model-free methods on long COT tasks significantly. VAPO not only demonstrates remarkable superiority in terms of performance but also showcases enhanced training efficiency, streamlining the learning process and underscoring its potential as a new benchmark in the field.
\item We propose Length-adaptive GAE, which adaptively adjusts the $\lambda$ parameter in GAE computation based on response lengths. By doing so, it effectively caters to the distinct bias-variance trade-off requirements associated with responses of highly variable lengths. As a result, it optimizes the accuracy and stability of the advantage estimation process, particularly in scenarios where the length of the data sequences varies widely.
\item We systematically integrate techniques from prior work, such as Clip-Higher and Token-level Loss from DAPO \citep{dapo}, Value-Pretraining and Decoupled-GAE from VC-PPO \citep{vc-ppo}, self-imitation learning from SIL \citep{SIL}, and Group-Sampling from GRPO \citep{shao2024deepseekmath}. Additionally, we further validate their necessity through ablation studies.
\end{enumerate}
 \textbf{VAPO} is an effective reinforcement learning system that brings together these improvements. These enhancements work together smoothly, leading to a combined result that’s better than the sum of the individual parts. 
 We conduct experiments using the Qwen2.5-32B pre-trained model, ensuring no SFT data is introduced in any of the experiments, to maintain comparability with related works (DAPO and DeepSeek-R1-Zero-Qwen-32B).
 The performance of \textbf{VAPO} improves from vanilla PPO a score of 5 to 60, surpassing the previous SOTA value-model-free methods DAPO~\cite{dapo} by 10 points. More importantly, \textbf{VAPO} is highly stable — we don't observe any crashes during training, and the results across multiple runs are consistently similar.

\section{Preliminaries}
This section presents the fundamental concepts and notations that serve as the basis for our proposed algorithm. We first explore the basic framework of representing language generation as a reinforcement learning task. Subsequently, we introduce Proximal Policy Optimization and Generalized Advantage Estimation.

\subsection{Modeling Language Generation as Token-Level MDP}
Reinforcement learning centers around the learning of a policy that maximizes the cumulative reward for an agent as it interacts with an environment. In this study, we cast language generation tasks within the framework of a Markov Decision Process (MDP) \citep{ouyang2022training}.

Let the prompt be denoted as $x$, and the response to this prompt as $y$. Both $x$ and $y$ can be decomposed into sequences of tokens. For example, the prompt $x$ can be expressed as $x=(x_0,\dots,x_m)$, where the tokens are drawn from a fixed discrete vocabulary $\mathcal{A}$.

We define the token-level MDP as the tuple $\mathcal{M}=(\mathcal{S},\mathcal{A},\mathbb{P},R,d_0,\omega)$. Here is a detailed breakdown of each component:
\begin{itemize}[leftmargin=*]
    \item \textbf{State Space ($\mathcal{S}$)}: This space encompasses all possible states formed by the tokens generated up to a given time step. At time step $t$, the state $s_t$ is defined as $s_t=(x_0,\dots,x_m,y_0,\dots,y_t)$.
    \item \textbf{Action Space ($\mathcal{A}$)}: It corresponds to the fixed discrete vocabulary, from which tokens are selected during the generation process.
    \item \textbf{Dynamics ($\mathbb{P}$)}: These represent a deterministic transition model between tokens. Given a state $s_t=(x_0,\dots,x_m,y_0,\dots,y_t)$, an action $a = y_{t + 1}$, and the subsequent state $s_{t+1}=(x_0,\dots,x_m,y_0,\dots,y_t,y_{t+1})$, the probability $\mathbb{P}(s_{t+1}|s_t,a)=1$.
    \item \textbf{Termination Condition}: The language generation process concludes when the terminal action $\omega$, typically the end-of-sentence token, is executed.
    \item \textbf{Reward Function ($R(s,a)$)}: This function offers scalar feedback to evaluate the agent's performance after taking action $a$ in state $s$. In the context of Reinforcement Learning from Human Feedback (RLHF)~\cite{rlhf,rlhf-sw}, the reward function can be learned from human preferences or defined by a set of rules specific to the task.
    \item \textbf{Initial State Distribution ($d_0$)}: It is a probability distribution over prompts $x$. An initial state $s_0$ consists of the tokens within the prompt $x$.
\end{itemize}

\subsection{RLHF Learning Objective}
We formulate the optimization problem as a KL-regularized RL task. Our objective is to approximate the optimal KL-regularized policy, which is given by:
\begin{align}\label{eq:objective}
   \pi^* =  \arg\max_\pi \mathbb{E}_{\pi, s_0 \sim d_0} \left[ \sum_{t = 0}^H  \left(R(s_t, a_t)-\beta \text{KL} \big( \pi(\cdot | s_t) \| \pi_{\text{ref}}(\cdot | s_t) \big)\right) \right]
\end{align}
In this equation, $H$ represents the total number of decision steps, $s_0$ is a prompt sampled from the dataset, $R(s_t, a_t)$ is the token-level reward obtained from the reward function, $\beta$ is a coefficient that controls the strength of the KL-regularization, and $\pi_{\text{ref}}$ is the initialization policy.

In traditional RLHF and most tasks related to LLMs, the reward is sparse and is only assigned at the terminal action $\omega$, that is, the end-of-sentence token \texttt{<eos>}. 

\subsection{Proximal Policy Optimization}
PPO \citep{ppo} uses a clipped surrogate objective to update the policy. The key idea is to limit the change in the policy during each update step, preventing large policy updates that could lead to instability.

Let $\pi_{\theta}(a|s)$ be the policy parameterized by $\theta$, and $\pi_{\theta_{\text{old}}}(a|s)$ be the old policy from the previous iteration. The surrogate objective function for PPO is defined as:

\begin{equation}
\mathcal{L}^{CLIP}(\theta)=\hat{\mathbb{E}}_t\left[\min\left(r_t(\theta)\hat{A}_t,\text{clip}(r_t(\theta), 1-\epsilon, 1+\epsilon)\hat{A}_t\right)\right]
\end{equation}

where $r_t(\theta)=\frac{\pi_{\theta}(a_t|s_t)}{\pi_{\theta_{\text{old}}}(a_t|s_t)}$ is the probability ratio, $\hat{A}_t$ is the estimated advantage at time step $t$, and $\epsilon$ is a hyperparameter that controls the clipping range.

Generalized Advantage Estimation \citep{schulman2015high} is a technique used to estimate the advantage function more accurately in PPO. It combines multiple-step bootstrapping to reduce the variance of the advantage estimates. For a trajectory of length $T$, the advantage estimate $\hat{A}_t$ at time step $t$ is computed as:

\begin{equation}
\hat{A}_t=\sum_{l = 0}^{T-t-1}(\gamma\lambda)^l\delta_{t + l}
\label{eq:gae_definition}
\end{equation}

where $\gamma$ is the discount factor, $\lambda\in[0, 1]$ is the GAE parameter, and $\delta_t=R(s_t, a_t)+\gamma V(s_{t + 1})-V(s_t)$ is the temporal-difference (TD) error. Here, $R(s_t, a_t)$ is the reward at time step $t$, and $V(s)$ is the value function. Since it is a common practice to use discount factor $\gamma = 1.0$ in RLHF, to simplify our notation, we omit $\gamma$ in later sections of this paper.

\section{Challenges in Long-CoT RL for Reasoning Tasks}
Long-CoT tasks present unique challenges to RL training, especially for methods that employ a value model to reduce variance. In this section, we systematically analyze the technical issues arising from sequence length dynamics, value function instability, and reward sparsity.

\subsection{Value Model Bias over Long Sequences}
\label{challenge1}
As identified in VC-PPO \citep{vc-ppo}, initializing the value model with a reward model introduces significant initialization bias. This positive bias arises from an objective mismatch between the two models. The reward model is trained to score on the \texttt{<EOS>} token, incentivizing it to assign lower scores to earlier tokens due to their incomplete context. In contrast, the value model estimates the expected cumulative reward for all tokens preceding \texttt{<EOS>} under a given policy. During early training phases, given the backward computation of GAE, there will be a positive bias at every timestep $t$ that accumulates along the trajectory.

Another standard practice of using GAE with $\lambda=0.95$ might exacerbates this issue. The reward signal $R(s_T, \text{<EOS>})$ at the termination token propagates backward as $\lambda^{T-t} R(s_T, \text{<EOS>})$ to the $t$-th token. For long sequences where $T-t \gg 1$, this discounting reduces the effective reward signal to near zero. Consequently, value updates become almost entirely bootstrapped, relying on highly biased estimates that undermine the value model's role as a reliable variance-reduction baseline.

\subsection{Heterogeneous Sequence Lengths during Training}

In complex reasoning tasks where a long CoT is essential for arriving at the correct answer, models often generate responses with highly variable lengths. This variability requires algorithms to be robust enough to manage sequences that can range from very short to extremely long. As a result, the commonly-applied GAE method with a fixed $\lambda$ parameter encounters significant challenges.

Even when the value model is perfect, a static $\lambda$ may not effectively adapt to sequences of varying lengths. For short-length responses, the estimates obtained through GAE tend to suffer from high variance. This is because GAE represents a trade-off between bias and variance. In the case of short responses, the estimates are skewed towards the variance-dominated side. On the other hand, for long-length responses, GAE often leads to high bias due to bootstrapping. The recursive nature of GAE, which relies on future state values, accumulates errors over long sequences, exacerbating the bias issue. These limitations are deeply rooted in the exponentially-decaying nature of GAE's computational framework.

\subsection{Sparsity of Reward Signal in Verifier-based Tasks}
\label{challenge3}

Complex reasoning tasks frequently deploy a verifier as a reward model \citep{o1, deepseekai2025deepseekr1incentivizingreasoningcapability}. Unlike traditional language-model-based reward models that provide a dense signal, such as a continuous value ranging from -4 to 4, verifier-based reward models typically offer binary feedback, such as 0 and 1. The sparsity of the reward signal is further compounded by long CoT reasoning. As CoT significantly elongates output lengths, it not only increases computational time but also reduces the frequency of receiving non-zero rewards. In policy optimization, the sampled responses with correct answer could be extremely scarce and valuable.

This situation poses a distinct exploration-exploitation dilemma. On one hand, the model must maintain relatively high uncertainty. This enables it to sample a diverse range of responses, increasing the likelihood of generating the correct answer for a given prompt. On the other hand, algorithms need to effectively utilize the correctly sampled responses—obtained through painstaking exploration—to enhance learning efficiency. By failing to strike the right balance between exploration and exploitation, the model may either get stuck in suboptimal solutions due to excessive exploitation or waste computational resources on unproductive exploration.
\section{VAPO: Addressing the Challenges in Long-CoT RL}

\subsection{Mitigating Value Model Bias over Long Sequences}
Building upon the analysis of value-model-based models presented in section \ref{challenge1}, we propose to use Value-Pretraining and decoupled-GAE to address the critical challenges in value model bias over long sequences. 
Both of these two techniques draw upon methodologies previously introduced in VC-PPO.

\textbf{Value-Pretraining} is proposed to mitigate the value initialization bias. Naively applying PPO to long-CoT tasks leads to failures such as collapsed output lengths and degraded performance. The reason is that the value model is initialized from the reward model while the reward model shares a mismatched objective with the value model. This phenomenon is first identified and addressed in VC-PPO \citep{vc-ppo}. In this paper, we follow the Value-Pretraining technique and the specific steps are outlined as follows:

\begin{enumerate}[leftmargin=*]
    \item Continuously generate responses by sampling from a fixed policy, for instance, $\pi_{\text{sft}}$, and update the value model with Monte-Carlo return.
    \item Train the value model until key training metrics, including value loss and explained variance \citep{explained_variance}, attain sufficiently low values.
    \item Save the value checkpoint and load this checkpoint for subsequent experiments. 
\end{enumerate}

\textbf{Decoupled-GAE} is proven effective in VC-PPO \citep{vc-ppo}. This technique decouples the advantage computation for the value and the policy. For value updates, it is recommended to compute the value-update target with $\lambda = 1.0$. This choice results in an unbiased gradient-descent optimization, effectively addressing the reward-decay issues in long CoT tasks.

However, for policy updates, using a smaller $\lambda$ is advisable to accelerate policy convergence under computational and time constraints. In VC-PPO, this is achieved by employing different coefficients in advantage computation: $\lambda_{\text{critic}} = 1.0$ and $\lambda_{\text{policy}} = 0.95$. In this paper, we adopt the core idea of decoupling GAE computation. 

\subsection{Managing Heterogeneous Sequence Lengths during Training}
To address the challenge of heterogeneous sequence lengths during training, we propose the \textbf{Length-Adaptive GAE}. This method dynamically adjusts the parameter in GAE according to the sequence length, enabling adaptive advantage estimation for sequences of varying lengths. Additionally, to enhance the training stability of mixed-length sequences, we replace the conventional sample-level policy gradient loss with a token-level policy gradient loss. The key technical details are elaborated as follows:

\textbf{Length-Adaptive GAE} is specifically proposed to to address the inconsistency in optimal $\lambda_{\text{policy}}$ values across sequences of varying lengths. In VC-PPO, $\lambda_{\text{policy}}$ is set to a constant value of $\lambda_{\text{policy}}=0.95$. However, when considering the GAE computation, for longer output sequences with lengths $l>100$, the coefficient of the TD-error corresponding to the reward is $0.95^{100}\approx0.006$, which is effectively zero. As a result, with a fixed $\lambda_{\text{policy}} = 0.95$, the GAE computation becomes dominated by potentially biased bootstrapping TD-errors. This approach may not be optimal for handling extremely long output sequences.

To address this shortcoming, we propose \textbf{Length-Adaptive GAE} for policy updates. Our method aims to ensure a more uniform distribution of TD-errors across both short and long sequences. We design the sum of the coefficients $\lambda_{\text{policy}}$ to be proportional to the output length $l$:
\begin{equation}
\sum_{t = 0}^{\infty}\lambda_{\text{policy}}^t\approx\frac{1}{1-\lambda_{\text{policy}}}=\alpha l,
\label{eq:variable_lam}
\end{equation}
where $\alpha$ is a hyper-parameter controlling the overall bias-variance trade-off. By solving Equation \ref{eq:variable_lam} for $\lambda_{\text{policy}}$, we derive a length-adaptive formula:
\begin{equation}
\lambda_{\text{policy}} = 1-\frac{1}{\alpha l}
\end{equation}

This length-adaptive approach to $\lambda_{\text{policy}}$ in GAE calculation allows for a more effective handling of sequences of varying lengths. 

\textbf{Token-Level Policy Gradient Loss}.
Following DAPO~\cite{dapo}, we have also modified the computation method of the policy gradient loss to adjust the loss weight allocation in long COT scenarios. Specifically, in previous implementations, the policy gradient loss was computed as follows:
\begin{align}
\mathcal{L}_{\text{PPO}}(\theta) =- \frac{1}{G} \sum_{i = 1}^G \frac{1}{|o_i|}\sum_{t = 1}^{|o_i|} \min \left(r_{i,t}(\theta) \hat{A}_{i,t}, \text{clip}\left(r_{i,t}(\theta), 1-\varepsilon, 1 + \varepsilon\right) \hat{A}_{i,t}\right),
\end{align}
where $G$ is the size of training batch, $o_i$ is the trajectory of the $i$th sample.
In this loss formulation, the losses of all tokens are first averaged at the sequence level before being further averaged at the batch level. This approach results in tokens from longer sequences contributing less to the final loss value. Consequently, if the model encounters critical issues in processing long sequences, a scenario that is prone to occur during the exploration phase of RL training, the insufficient suppression caused by their diminished weighting may lead to training instability or even collapse. To address this imbalance in token-level contribution to the final loss, we revise the loss function into the following form:
\begin{align}
\mathcal{L}_{\text{PPO}}(\theta) =- \frac{1}{\sum_{i = 1}^G |o_i|} \sum_{i = 1}^G \sum_{t = 1}^{|o_i|} \min \left(r_{i,t}(\theta) \hat{A}_{i,t}, \text{clip}\left(r_{i,t}(\theta), 1-\varepsilon, 1 + \varepsilon\right) \hat{A}_{i,t}\right),
\end{align}
where all tokens within a single training batch are assigned uniform weights, thereby enabling the problems posed by long sequences to be addressed with enhanced efficiency.

\subsection{Dealing with Sparsity of Reward Signal in Verifier-based Tasks}
As analyzed in Section \ref{challenge3}, enhancing the efficiency of exploration-exploitation tradeoff in RL training becomes critically challenging under scenarios with highly sparse reward signals. To address this key issue, we adopt three methods: Clip-Higher, Positive Example LM Loss and Group-Sampling. The technical details are elaborated as follows:

\textbf{Clip-Higher} is used to mitigate the entropy collapse issue encountered in PPO and GRPO training process, which is first proposed in DAPO \citep{dapo}. We decouple the lower and higher clipping range as $\varepsilon_\text{low}$ and $\varepsilon_\text{high}$
\begin{align}
\mathcal{L}_{\text{PPO}}(\theta) =- \frac{1}{\sum_{i = 1}^G |o_i|} \sum_{i = 1}^G \sum_{t = 1}^{|o_i|} \min \left(r_{i,t}(\theta) \hat{A}_{i,t}, \text{clip}\left(r_{i,t}(\theta), 1-\textcolor{red}{\varepsilon_\text{low}}, 1 + \textcolor{red}{\varepsilon_\text{high}}\right) \hat{A}_{i,t}\right),
\end{align}
We increase the value of $\varepsilon_\text{high}$ to leave more room for the increase of low-probability tokens. We opt to keep $\varepsilon_\text{low}$ relatively small, because increasing it will suppress the probability of these tokens to 0, resulting in the collapse of the sampling space.

\textbf{Positive Example LM Loss} is designed to enhance the utilization efficiency of positive samples during RL training process. In the context of RL for complex reasoning tasks, some tasks demonstrate remarkably low accuracy, with the majority of training samples yielding incorrect answers. Traditional policy optimization strategies that suppress the generation probability of erroneous samples suffer from inefficiency during RL training, as the trial-and-error mechanism incurs substantial computational costs. Given this challenge, it is critical to maximize the utility of correct answers when they are sampled by the policy model. To address this challenge, we adopt an imitation learning approach by incorporating an additional negative log-likelihood (NLL) loss for the correct outcomes sampled during RL training. The corresponding formula is as follows:
\begin{align}
\mathcal{L}_{\text{NLL}}(\theta) =- \frac{1}{\sum_{o_i \in \mathcal{T}}|o_i|} \sum_{o_i \in \mathcal{T}} \sum_{t = 1}^{|o_i|} \log\pi_{\theta}\left(a_t|s_t\right),
\end{align}
where $\mathcal{T}$ denotes the set of correct answers. The final NLL loss is combined with the policy gradient loss through a weighting coefficient $\mu$, which collectively serves as the objective for updating the policy model:
\begin{align}
\mathcal{L}(\theta) = \mathcal{L}_{\text{PPO}}(\theta) + \mu * \mathcal{L}_{\text{NLL}}(\theta).
\end{align}

\textbf{Group-Sampling} is used to sample discriminative positive and negative samples within the same prompt. 
Given a fixed computational budget, there exist two primary approaches to allocating computational resources. The first approach utilizes as many prompts as possible, with each prompt sampled only once. The second approach reduces the number of distinct prompts per batch and redirects computational resources toward repeated generations. We observed that the latter approach yields marginally better performance, attributed to the richer contrastive signals it introduces, which enhance the policy model’s learning capability.
\section{Experiments}

\subsection{Training Details}
In this work we enhanced the model's mathematical performance by introducing various modifications to the PPO algorithm based on the Qwen-32B model. These techniques are also effective for other reasoning tasks, such as code-related tasks. For the basic PPO, we used AdamW as the optimizer, setting the actor learning rate to \(1 \times 10^{-6}\) and the critic learning rate to \(2 \times 10^{-6}\), as the critic needs to update faster to keep pace with policy changes. The learning rate employed a warmup-constant scheduler. The batch size was 8192 prompts, with each prompt sampled once, and each mini-batch size set to 512. The value network was initialized using a reward model, with the GAE $\lambda$ set to 0.95 and $\gamma$ set to 1.0. Sample-level loss was used, and the clip $\epsilon$ was set to 0.2.

Compared to vanilla PPO, VAPO made the following parameter adjustments:
\begin{enumerate}
\item Implemented a value network warmup for 50 steps based on the reward model (RM) before initiating policy training.
\item Utilized decoupled GAE, where the value network learns from returns estimated with $\lambda$=1.0, while the policy network learns from advantages obtained using a separate lambda.
\item Adaptively set the lambda for advantage estimation based on sequence length, following the formula: $\lambda_{\text{policy}} = 1-\frac{1}{\alpha l}$, where $\alpha=0.05$.
\item Adjusted the clip range to $\epsilon_\text{high}$=0.28 and $\epsilon_
\text{low}$=0.2.
\item Employed token-level policy gradient loss.
\item Added a positive-example language model (LM) loss to the policy gradient loss, with a weight of 0.1.
\item Used 512 prompts per sampling, with each prompt sampled 16 times, and set the mini-batch size to 512.
\end{enumerate}

We will also demonstrate the final effects of removing each of these seven modifications from VAPO individually. For the evaluation metric, we use the average pass rate of AIME24 over 32 times, with sampling parameters set to topp=0.7 and temperature=1.0.

\subsection{Ablation Results}

\setcounter{table}{0}
\begin{table}[t]
    \centering
    \caption{Abalation results of \textbf{VAPO}}
    \begin{tabular}{l c}
        \toprule
        \textbf{Model} & $\textbf{AIME24}_\text{avg@32}$ \\
        \midrule
        Vanilla PPO & 5 \\
        \textbf{DeepSeek-R1-Zero-Qwen-32B}  & 47 \\
        \textbf{DAPO} & 50 \\
        \midrule
        VAPO w/o Value-Pretraining & 11 \\
        VAPO w/o Decoupled-GAE & 33 \\
        VAPO w/o Length-adaptive GAE & 45 \\
        VAPO w/o Clip-Higher & 46 \\
        VAPO w/o Token-level Loss & 53 \\
        VAPO w/o Positive Example LM Loss & 54 \\
        VAPO w/o Group-Sampling & 55 \\
        \textbf{VAPO} & \textbf{60} \\
        \bottomrule
    \end{tabular}
    \label{tab:results}
\end{table}

On Qwen-32b, DeepSeek R1 using GRPO achieves 47 points on AIME24, while DAPO reaches 50 points with 50\% of the update steps. In Figure~\ref{fig:front}, our proposed VAPO matches this performance using only 60\% of DAPO's steps and achieves a new SOTA score of 60.4 within just 5,000 steps, demonstrating VAPO's efficiency. Additionally, VAPO maintains stable entropy—neither collapsing nor becoming excessively high—and consistently achieves peak scores of 60-61 across three repeated experiments, highlighting the reliability of our algorithm.

\Cref{tab:results} systematically presents our experimental results. The Vanilla PPO method, hindered by value model learning collapse, only achieves 5 points in the later stages of training, characterized by a drastic reduction in response length and the model directly answering questions without reasoning.
Our VAPO method finally achieves 60 points, which is a significant improvement. We further validated the effectiveness of the seven proposed modifications by ablating them individually:
\begin{enumerate}
\item Without Value-Pretraining, the model experiences the same collapse as Vanilla PPO during training, converging to a maximum of approximately 11 points.
\item Removing the decoupled GAE causes reward signals to exponentially decay during backpropagation, preventing the model from fully optimizing long-form responses and leading to a 27-point drop.
\item Adaptive GAE balances optimization for both short and long responses, yielding a 15-point improvement.
\item Clip higher encourages thorough exploration and exploitation; its removal limited the model's maximum convergence to 46 points.
\item Token-level loss implicitly increased the weight of long responses, contributing to a 7-point gain.
\item Incorporating positive-example LM loss boosted the model by nearly 6 points.
\item Using Group-Sampling to generate fewer prompts but with more repetitions also resulted in a 5-point improvement.
\end{enumerate}

\subsection{Training Dynamics}

\begin{figure}[t]
    \centering
    \begin{subfigure}{0.45\textwidth}
        \centering
        \includegraphics[width=\textwidth]{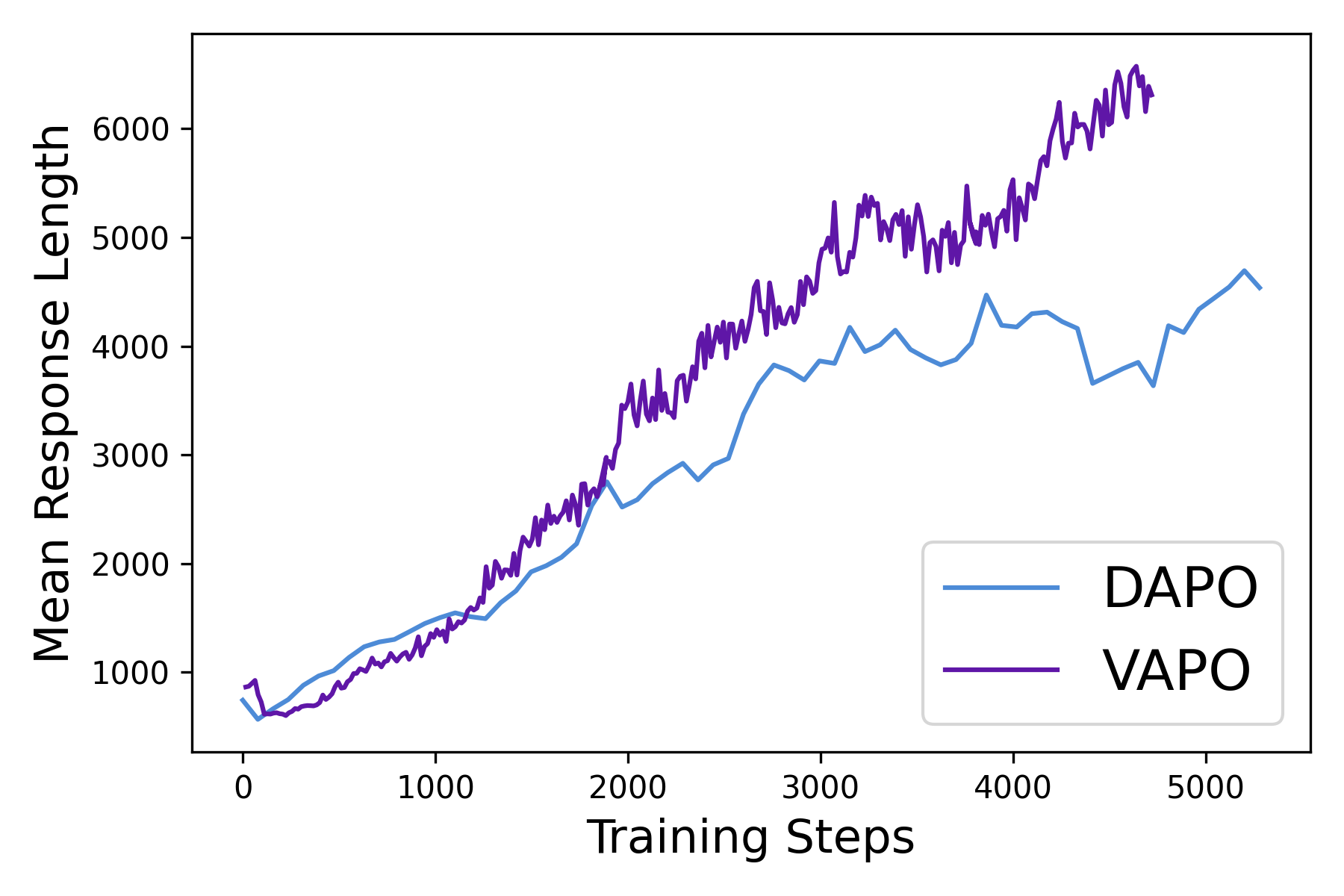}
        \caption{Mean response length.}
        \label{subfig:length}
    \end{subfigure}
    \begin{subfigure}{0.45\textwidth}
        \centering
        \includegraphics[width=\textwidth]{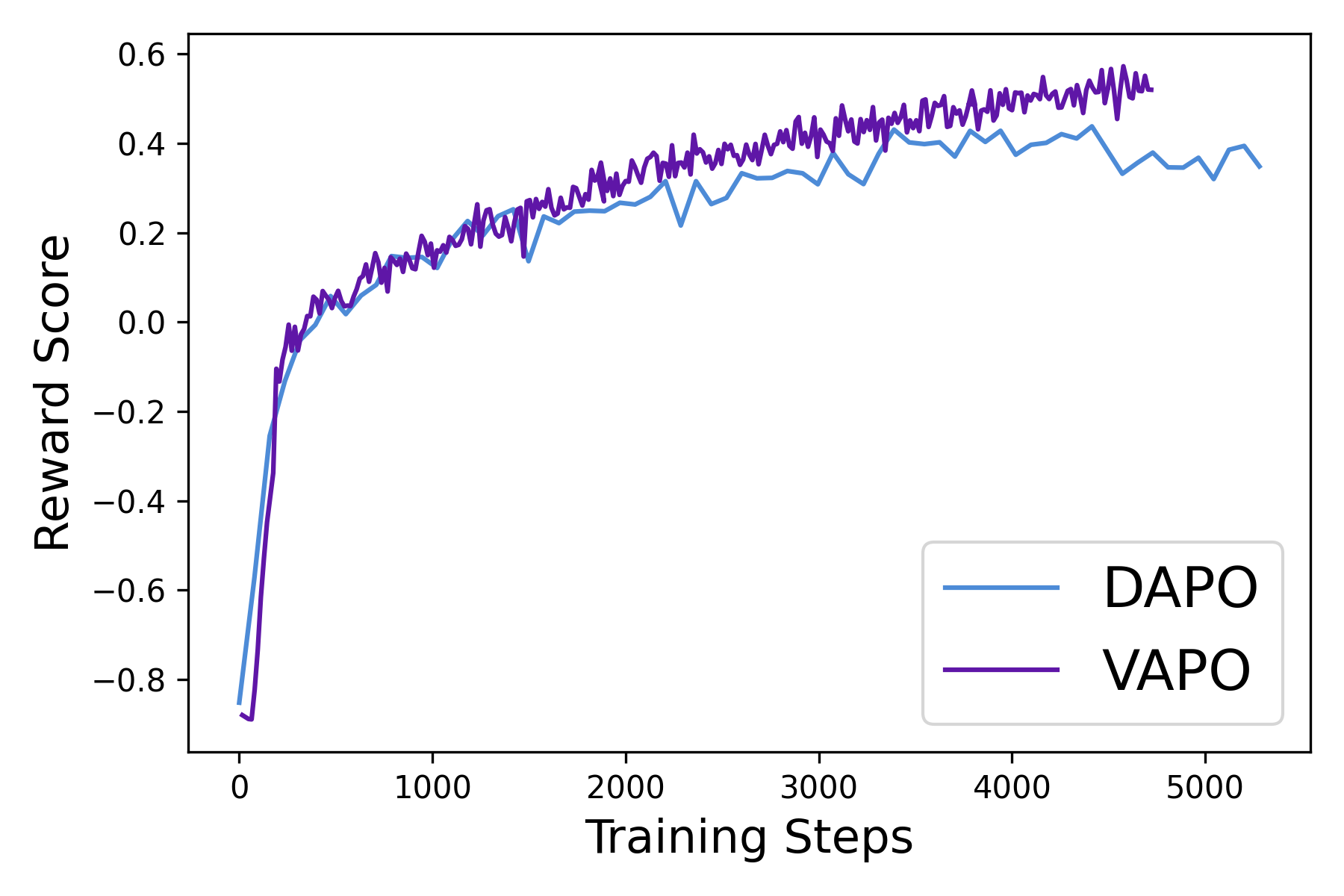}
        \caption{Reward score.}
        \label{subfig:reward}
    \end{subfigure}
    \begin{subfigure}{0.45\textwidth}
        \centering
        \includegraphics[width=\textwidth]{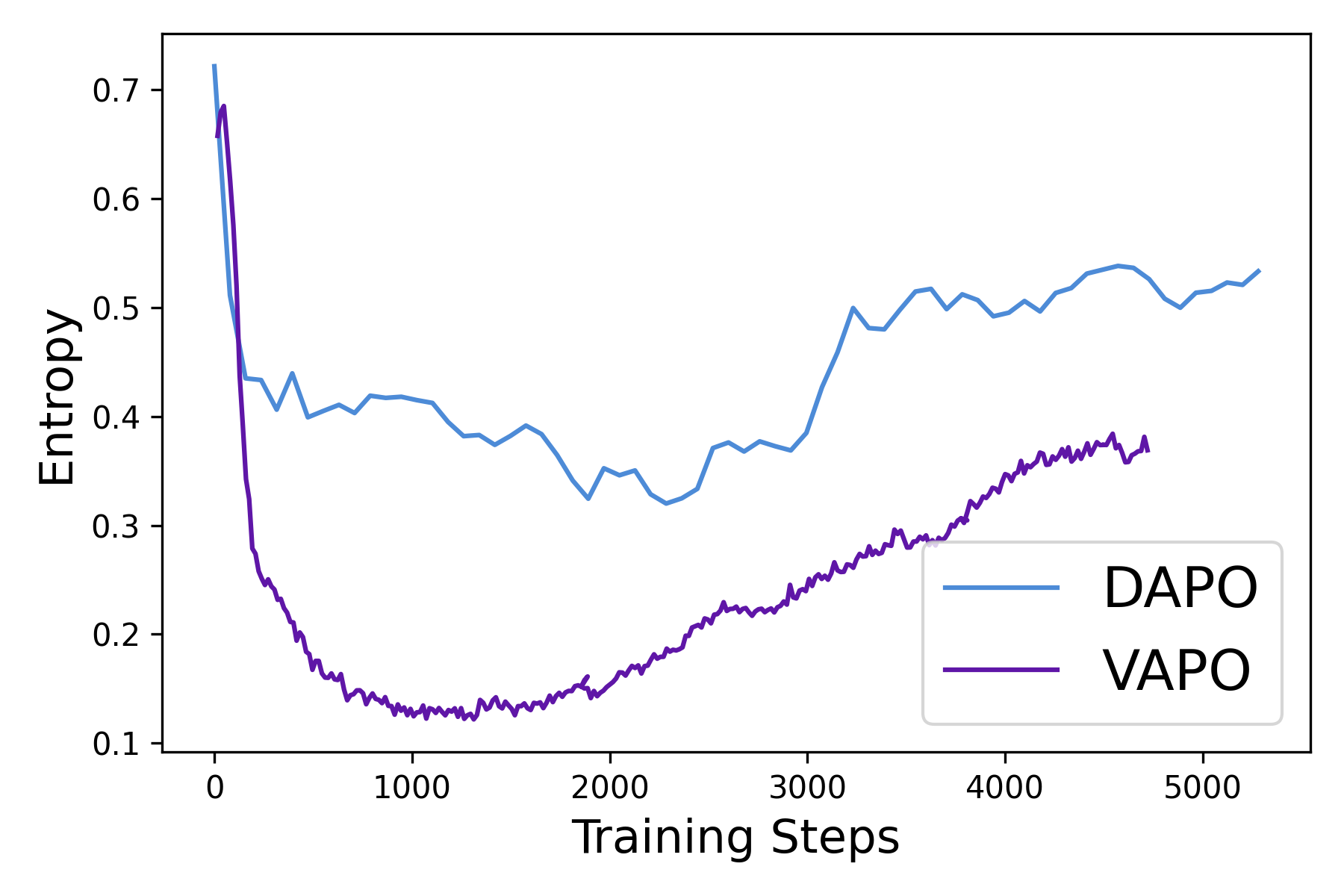}
        \caption{Generation entropy.}
        \label{subfig:entropy}
    \end{subfigure}
    \caption{VAPO's metric curves for response length, reward score, and generation entropy.}
    \label{fig:metrics}
\end{figure}

The curves generated during RL training provide real-time insights into training stability, and comparisons between different curves can highlight algorithmic differences. It is generally believed that smoother changes and faster growth are the desirable characteristics of these curves. Through a comparison of the training processes of VAPO and DAPO, we made the following observations:
\begin{itemize}
\item \Cref{fig:metrics} shows that VAPO's training curve is smoother than DAPO's, indicating more stable algorithmic optimization in VAPO.
\item As depicted in \Cref{subfig:length}, VAPO exhibits superior length scaling compared to DAPO. In modern contexts, better length scaling is widely recognized as a marker of improved model performance, as it enhances the model's generalization capabilities.
\item \Cref{subfig:reward} demonstrates that VAPO's score grows faster than DAPO's, as the value model provides the model with more granular signals to accelerate optimization.
\item According to \Cref{subfig:entropy}, VAPO's entropy drops lower than DAPO's in the later stages of training. This is two sides of the coin: on one hand, it may hinder exploration, but on the other hand, it improves the model stability. From VAPO's final results, the lower entropy has minimal negative impact on performance, while the reproducibility and stability proves highly advantageous.
\end{itemize}

\section{Related Work}

OpenAI O1 \citep{o1} introduces a profound paradigm shift in LLMs, characterized by extended reasoning before delivering a final response~\cite{grok,qwq,gemini-thinking}. 
DeepSeek R1 \citep{deepseekai2025deepseekr1incentivizingreasoningcapability} open-sources both its training algorithm (the value-model-free GRPO \cite{shao2024deepseekmath}) and its model weights, which are comparable in performance to O1. DAPO \cite{dapo} identifies previously undisclosed challenges such as entropy collapse encountered during the scaling of value-model-free LLM RL, and proposes four effective techniques to overcome these challenges, achieving SOTA industry-level performance. Recently, Dr. GRPO \cite{liu2025understandingr1zeroliketrainingcritical} removes both the length and std normalization terms in GRPO.
On the other hand, ORZ \cite{hu2025openreasonerzeroopensourceapproach} follows PPO and utilizes a value model for advantage estimation, proposing Monte Carlo estimation instead of Generalized Advantage Estimation. However, they could just achieves a comparable performance to value-model-free method like GRPO and DAPO. In this paper, we also follow the value-model-based approach and propose VAPO, which outperforms the SOTA value-model-free algorithm DAPO.

\section{Conclusion}
In this paper, we propose an algorithm named VAPO, which leveraging the Qwen2.5-32B model, achieves the SOTA performance on the AIME24 benchmark. By introducing seven novel techniques atop PPO, which focus on refining value learning and balancing exploration, our value-model-based approach outperforms contemporary value-model-free methods like GRPO and DAPO. The work provides a robust framework for advancing large language models in reasoning-intensive tasks.

\newpage

\section*{Contributions}

\textbf{Project Lead}\quad 

Yu Yue$^{1}$


\textbf{Algorithm}

Yu Yue$^{1}$, Yufeng Yuan$^{1}$, Qiying Yu$^{1,2}$, Xiaochen Zuo$^{1}$, Ruofei Zhu$^{1}$, Wenyuan Xu$^{1}$, Jiaze Chen$^{1}$, Chengyi Wang$^{1}$, TianTian Fan$^{1}$,  Zhengyin Du$^{1}$, Xiangpeng Wei$^{1}$, Xiangyu Yu$^{1}$


\textbf{Infrastructure$^{*}$}

Gaohong Liu$^{1}$, Juncai Liu$^{1}$, Lingjun Liu$^{1}$, Haibin Lin$^{1}$, Zhiqi Lin$^{1}$, Bole Ma$^{1}$, Chi Zhang$^{1}$, Mofan Zhang$^{1}$, Wang Zhang$^{1}$, Hang Zhu$^{1}$, Ru Zhang$^{1}$

$^{*}$Last-Name in Alphabetical Order



\textbf{Supervision}

Xin Liu$^{1}$, Mingxuan Wang$^{1}$, Yonghui Wu$^{1}$, Lin Yan$^{1}$


\textbf{Affiliation}

$^1$ ByteDance Seed

$^2$ SIA-Lab of Tsinghua AIR and ByteDance Seed

\clearpage

\bibliographystyle{plainnat}
\bibliography{main}


\end{document}